# Review: Modeling and Classical Controller Of Quad-rotor

Tarek N.Dief
Department of Earth System Science and Engineering
Kyushu University
Fukuoka, Japan

Shigeo Yoshida
Research Institute for Applied Mechanics
Kyushu University
Fukuoka, Japan

*Abstract*—**This paper presents an overview of the most effective ideas for the Quad-rotor project. The concept of modeling using different methods is presented. The modeling part presented the nonlinear model, and the concept of linearization using small disturbance theory. Parameter identifications part explained the most important parameters that affect the system stability and tried to get suitable solutions for these problems and identify some parameters experimentally. Data filtration, Kalman filter, Structure design, motor distribution, aerodynamic effect, analysis of shroud and its effect on the resultant thrust were explained. The control part incorporates different classical schemes such as PD and PID controllers to stabilize the Quad-rotor. Also, different ideas are presented to stabilize the quad rotor using PID controllers with some modification to get high maneuverability and better performance.**

## I.  INTRODUCTION

Nowadays, automatic flying of intelligent vehicles represents a huge field of applications. The rapid development of microcontroller affects all control applications which mean new control theory, new researches, and new challenges. The application of the hovering aerial vehicles gave chances for researchers to present better controllers to achieve, the more aggressive mission for the Quad-rotor. In the beginning of this century, the chance to apply new control algorithms to small aircraft increased rapidly. The rapid developments in applications are also due to the advances in computation systems, communication systems, motors, and sensors which are getting very cheaper with to higher performance. Using new microcontrollers will give Quad-rotor chance to be smaller with higher performance. Nonlinear control theories and difficult applications are the target of the researchers to present and develop more controllers that were very difficult to be implemented using microcontrollers with higher sample rate.  Some of these nonlinear controllers are back step controller, and sliding mode controller. Also the optimal control is presented, especially LQR (Linear Quadratic Regulator) and LQG (Linear Quadratic Gaussian) controllers to stabilize the Quad-rotor. Classic controllers are widely used to stabilize the Quad-rotor with some modification to be suitable for all systems and could give smoother responses with lower sample rate.

This paper focused on the classic controller for its simplicity in implementation and tuning.

However, the robustness of the classic controller are low, so some modifications are presented to these controllers to be more robust. The most important works in classic controller are applied with small modifications to stabilize the Quad-rotor [1-5].

There are a lot of applications for the Quad-rotor such as line tracking [6-8], photo capturing, pollution measuring, and high endurance flight using fixed wing to reduce the power consuming [9].

Line tracking is one of the most important applications for the Quad-rotor. A lot of trials presented to simulate this idea and to get successful results [6-8]. The concept of line tracking is to make the system autonomous, and give it certain track and make the system follows this track. But the application for this system will be limited to indoor, as it need sensors to get a position as a feedback with high accuracy.

Power saving is one of the applications for Quad-rotor using classical control [9]. Fixed wing is used to generate force as a thrust, and lift. This concept will make the mathematical model of the Quad-rotor changed, and the condition of symmetry will be violated.

The fault that may occur to the brushless motor [10] will make the system failure. So using alternative ideas to control the system even any failure occurs to the motors, also this paper presented the idea and control algorithm to control the system for the new case.

## II.  MODELING AND LINEARIZTION

### A.  General:

For studying the mathematical model of the Quad-rotor, we need to make some assumptions for system dynamics. Some of these assumptions are as below.

- Quad-rotor's structure is symmetric around.

- Quad-rotor UAV consists of a rigid frame equipped with four rotors.

- No mass changes during motion (time invariant).

- COG (Center of Gravity) is fixed at the origin of the Quad-rotor.

- Thrust and drag constants are proportional to the square value of the motor's speed.

There are two methods for deriving the system equation. These two methods depend on transformation matrix which describes the relative motion between body axis and inertial axis [11-19].

The first method is Newton-Euler method; this method depends on the rigid body equations that result from Newton's second law [11,20]. The second method is Euler-Lagrange; this method depends on the energy and kinematics to get the mathematical model [12,19]. But these two techniques should give the same results.

There is another term comes from the gyroscopic effect due to the high RPM for motors, but it will be vanished during the linearization [11].There are some techniques to rearrange the rotor of the Quad-rotor; but these techniques will be more complex than the normal case of Quad-rotor at modelling and control design [21], some of these techniques use different sizes of rotors in the same system which mean different thrust and drag, and also some of them fix the rotors in different distances from the origin which means that the assumptions of symmetry will be violated. Another technique is used to make one rotor of the four rotor change the direction by 90 degrees to get axial motion only [16], this technique give a chance for the system to get fast motion along one axis. Also, it increases the maneuverability of the system.

### B. Nonlinearlity and multi-imput:

The mathematical model of the Quad-rotor is nonlinear; so the classical control can't be used directly without transforming this mathematical model from time domain to s-domain.

As a condition of the transformation, linear and time-invariant model is necessary. Some methods are applied to linearize the nonlinear model; one of the most popular methods is the small disturbance theory [11, 22] to linearize the system.

The Quad-rotor system is MIMO (multi-input multi-output); which means that it is a function of the four rotors. So we need to get a technique that makes the system SISO (single-input single- output). Some ideas were suggested to get just one input for each output [20, 23]. This technique may not be completely correct as the four rotors don't have the same performance, but it has a small effect and this effect can be solved by the controller action. After using this technique, we can get the output such as the altitude as a function of single input [11].

### C. Rotor and motor identification:

There are a lot of projects tried to design the blade of the Quad-rotor. Some of these projects gave an analytical estimation for the system parameters. And other projects gave a numerical solution for the rotor performance using simulation programs [24].

There is other type of the projects tried to get the rotor parameters by experimental methods; this technique depends on changing the RPM's values and get the corresponding thrust and drag. By changing the rotor's RPM we can get a linear relation between the square values of the angular velocity versus thrust and drag [25-27]. Some techniques used duct or shroud; the target from these techniques is to get higher thrust [28-29]. Also, some projects used this technique for safety.

### D. Saturation:

For the Quad-rotor system, the control action will be limited by the specifications of the motors. So give action to the motors over the maximum limit may cause a failure to the motors. So limits are needed in the controller (saturation). The saturation for motor action will be considered as a nonlinear term, also it will have a great effect in the system response [30, 31]; the settling time for the system will be longer than the normal case without saturation. So during designing the control system, we have to consider this effect in simulation before implementing it in the Quad-rotor platform.

### E. Sample time:

There are two methods for controlling the Quad-rotor. The first method is to make the system full autonomous to achieve the mission without any interrupt from another system like ground station during flight. This method is applied for some systems such as the line tracking; the required mission is provided on the system microcontroller. Feedback will be provided to the microcontroller automatically from sensors, after that the controller will take the suitable action to achieve the required mission. In this case the controller code is already written on the microcontroller.

The other method will be the same like the first method except the situation of the controller code. The controller code will be stored on ground station and the Quad-rotor will send the feedback to the ground station and receive the control action from the ground station.

For both of these two methods, the sample time will be critical for the system stability. There will be a certain limit for the sample time to stabilize the Quad-rotor; this limit will be function of the Quad-rotor size. The size of the Quad-rotor will increase proportionally with the maximum limit of the sample time.

In the normal case of the small Quad-rotor, which has diameter size around 50 centimeters, the sample time limit will be around 0.025 second [32]. The sample time also depends on the system hardware. So we need to choose microcontroller suitable for the system.

### F. Motor dead zone & calibration:

Brushless motor has a small non effective region in the low values of the Pulse Width Modulation (PWM), this region is called dead zone. Dead zone is the area that the motor receives order to rotate, but it generates no rotation. This region is not constant and it depends on motors and speed controllers.

At the first using of any brushless motors, you have to make calibration. Calibration will give the maximum and minimum values of the PWM to the speed controller.

### G. Filtration of sensor data:

Using the filtration for the system data was very common with researchers who are working on robots and autonomous systems such as Quad-rotor. They were using the data filtration as the sensor data had a lot of noise. This noise can make the system response unstable. One of the most common filters is Kalman filter [33-35], also there are other filters applied to the system, these filters depend on the accuracy of the sensors.

Now, sensor data are better and we can use it without filtration, as the filtration algorithm is already embedded on the sensor chip.

### H. Sensor data:

Feedback is used to design the controller; also the accuracy of sensor data has a great effect on system stability. IMU (Inertial Measurement Unit) is used as a feedback to get system angles. Now, IMU is very cheap and gives very accurate readings. Some projects can use Kalman filter [35] for more accuracy, also some projects use controllers with a combined with Kalman filter such as LQG controller.

To get position, it is complex to get the position directly during outdoor flight such as IMU. We can use different types of sensors to get altitude; one of these sensors is sonar. Sonar can give very high accurate readings, but the range of sonars is very small, it can reach to 6 meters only. Also, there are some sensors like altimeters which can give readings in high altitudes, but the accuracy of the altimeter is as small as 50 centimeters.

There are other types of sensors that can get the position. One of these sensors is global positioning system (GPS). Commercial GPS has a very low accuracy which can reach to 6 meters. In that case this accuracy will lead to instability for the Quad-rotor. One the other hand, there is GPS can get very high accuracy, but it's used in military applications and not available for all researchers. So for outdoor flight, researchers just use sonar with altimeter to control the altitude only. There are other techniques used for the indoor flight to get altitude and position by a set of cameras. These cameras are fixed in different points, then we can make some simple calculations to get the system position [8, 36-41]. One of the most important techniques that depend on this idea is called Moiré technique [42-43].

### III. CLASSICAL CONTROL

The open loop transfer function of the Quad-rotor mainly consists of double integrators; also we have another pole as a motor delay. For the motor simulation, there were a lot of experiments had done to get a mathematical model. The mathematical model for the motor transfer function is a first order system with a delay of 0.1 second [12, 27, 31].

But the motor delay will decay faster than the double integrator and the double integrator effect will be the dominant in the system dynamics [6, 11, 27,31].

In designing controller for the Quad-rotor there are some steps important for nonlinear systems. So we can apply any controller to the linear model, and then apply the same controller to the real Quad-rotor platform. Sometimes after applying the controller to the real platform, we found that the system is unstable due to the effect of the linearization. So testing the controller on the linear and nonlinear simulation is very important before using it on the real model.

### A. PD controller:

For this kind of controller, it can give good results in the ideal case without any disturbance. But for the real case, we will have a problem of the steady state error. Designing the controller depends on the error and the error rate only. So the control action responses due to the effect of big errors and fast disturbance. The steady state error will be very critical, as the steady state error for the Quad-rotor will be around 2 degrees. This means that high drift will happen for the Quad-rotor [23, 44].

### B. PID controller:

The concept of the PID is the same like the PD controller, but there is another term added to control the effect of the steady state error. Cancelling the effect of the steady state error will be represented by adding an integrator term to the controller. To apply this controller will need to get old data as a history and try to restore the system due to this history. This integrator term will take action in case of resultant error is not equal zero. On the other hand, adding the integrator term to the PID controller will lead to add an integrator to the system. So it increases the chance of the instability in the system. So tuning the system will be more critical [11, 14, 27, 45-47].

### C. Improved techniques related to PID controller:

In this paper [48], different approaches of the PID controller are presented. The concept of the PID consists of three terms, proportional, derivative and integrator. The ideal case for the PID controller is illustrated in this paper, after that the structure of the three terms is changed with different signs. Also three approaches are used based on the PID controller concept to present the idea of the controllers. The performance will be varied as a result of changing the order of the block diagram. Finally, comparison between the three different approaches and their responses were discussed in details.

In this paper [6], the concept of the PID controller is applied, but there is another term is added for aggressive maneuver. This controller consists of four terms. The first three terms are the same like the ideal PID controller. And the fourth term is trying to stabilize the rate of the Euler angle rate (acceleration) and to overcome the high acceleration effect.

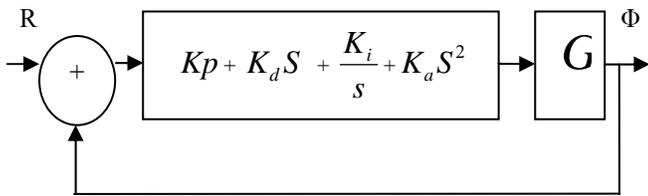

Figure 1.block diagram m of PID controller
with acceleration term

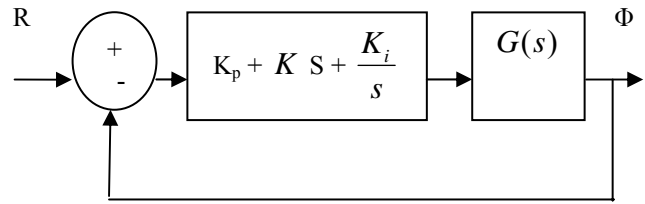

Figure 2. block diagram of PID controller

In this paper [49], a new technique for the PID controller is presented based on the genetic algorithm GA. genetic algorithm and improved genetic algorithm are applied for tuning PID controller's gains. The design steps of the controller explained the comparison between genetic algorithms and the improved genetic algorithm. Also the concept of Ziegler-Nichols method is presented and implemented in the Quad-rotor case. For the improved genetic algorithm, it can achieve a global optimal solution for the cost function with less settling time compared with the ideal case of PID controller.

In this paper [50], the fuzzy controller is applied to the system to just to get the PID controller gains which stabilize the Quad-rotor.

This technique applied the same steps for the fuzzy controller; it considered the error and rate of error as input after that makes the steps of the design, fuzzification, rule base, and defuzzification. So the resultant outputs from fuzzy algorithm are the gains of the PID controller. The results of this technique show that the settling time for the combined fuzzy-PID controller is 0.6 second, and the ideal case of the PID controller is 1.5 second.

This paper used seven memberships for the error and rate of error. It will give a very high accurate response in the shorter settling time. But the problem will be in the implementation. This controller is applied to the simulation only, but implementing this controller to the real model will be more difficult and needs hardware with high frequencies to achieve the same response.

In this paper [51], a technique called iterative learning control (ILC) was used. This technique applied an iterative method to get the suitable control action. The combination of the classic controller and previous control action are combined with the convergence condition for stability. Three different approaches were applied to control the Quad-rotor system using this technique; offline ILC, online ILC, and online-offline ILC. All these techniques depend on the iteration and satisfying the convergence condition. These techniques are very simple for controlling the Quad-rotor and implementation. Applying this controller will increase the robustness of the system and also overcome any disturbance. Also; it gives response with sample time longer than the ideal PID controller.

In this paper [40], controlling the Quad-rotor is the same like [11], but the mathematical model replaced with another digital system to get all the .system states using a vanishing point algorithm. This algorithm can transform group on parallel lines in the three-dimension (3D) space into some intersection lines in two-dimension (2D) image, and the intersection between these lines will be the vanishing point

This algorithm is valid for indoor flight only and can give a complete definition of the Quad-rotor parameters.

After estimating the system model, the PID controller is applied based on the digital model to stabilize the Quad-rotor system.

In this paper [52], the new technique is presented to make a combination of PID and particle swarm optimization (PSO) method. This technique used to find the parameters required to optimize a particular object. Applying this technique can get the minimum and maximum values for parameters such as cost function.

Also, another technique called as a bacterial foraging optimization (BFO) is used. This technique also depends on the optimization to get the system parameters. The last technique used in this paper is mixing between the two techniques and it is called (PSO-BFO).

A comparison between different control algorithms for the Quad-rotor is presented [52], and the advantages and disadvantages for each controller. So choosing your controller will depend on the mission for your Quad-rotor and the Hardware which you are using in the system

IV. CONCLUSION

This model solve a very important problem we met during tests the quad rotor controller which is the asymmetry that results from the motors; as each motor has a little different responses with the time. So the update of the parameter estimation in our model can put the change in motor parameters under consideration. The Quad rotor is nonlinear problem, so the initial conditions and the input will play a huge role in the control design; so using adaptive control will make the control adapt with any problem or change occur in the system. By comparing the response between the classic controller like PID and the adaptive control we will find that the contribution of nonlinearity high during using PID controller as it depends on the system model which we assume it's constant, on the other hand the adaptive will be more reliable in different cases for the flight motion and ready for any change or disturbance occur to the system and also put it as a part of the model..

REFERENCES

[1] D. Mellinger, M. Shomin, and V. Kumar, "Control of Quadrotors for Robust Perching and Landing.pdf," *Int. Powered Lift Conf.*, pp. 119–126, 2010.

[2] M. Hehn and R. D'Andrea, "A flying inverted pendulum," *Proc. - IEEE Int. Conf. Robot. Autom.*, no. 2, pp. 763–770, 2011.

[3] S. Fowers, "Stabilization and Control of a Quad-rotor Micro-UAV Using Vision Sensors," *Eng. Technol.*, no. August, 2008.

[4] J. H. Gillula, H. Huang, M. P. Vitus, and C. J. Tomlin, "Design of guaranteed safe maneuvers using reachable sets: Autonomous quadrotor aerobatics in theory and practice," *Proc. - IEEE Int. Conf. Robot. Autom.*, pp. 1649–1654, 2010.

[5] Seul Jung, "A Position-Based Force Control Approach to a Quad-rotor System," *Ubiquitous Robots and Ambient Intelligence (URAI), 2012 9th International Conference on* , vol., no., pp.373,377, 26-28 Nov. 2012

[6] G. M. Hoffmann, S. L. Waslander, and C. J. Tomlin, "Quadrotor Helicopter Trajectory Tracking Control," *Electr. Eng.*, vol. 44, no. August, pp. 1–14, 2008.

[7] Guney, M.A.; Unel, M., "Formation Control of a Group of Micro Aerial Vehicles (MAVs)," *Systems, Man, and Cybernetics (SMC), 2013 IEEE International Conference on* , vol., no., pp.929,934, 13-16 Oct. 2013

[8] S. Shen, Y. Mulgaonkar, N. Michael, and V. Kumar, "Vision-based state estimation and trajectory control towards high-speed flight with a quadrotor," *Robot. Sci. ...* , 2013.

[9] Oosedo, A.; Konno, A.; Matumoto, T.; Go, K.; Masuko, K.; Abiko, S.; Uchiyama, M., "Design and simulation of a quad rotor tail-sitter unmanned aerial vehicle," *System Integration (SII), 2010 IEEE/SICE International Symposium on* , vol., no., pp.254,259, 21-22 Dec. 2010

[10] A. B. Milhim, Y. Zhang, C. Rabbath, M. Student, and D. Scientist, "Gain Scheduling Based PID Controller for Fault Tolerant Control of a Quad-Rotor UAV," *Flying*, no. April, pp. 1–13, 2010.

[11] H. Bolandi, M. Rezaei, and R. Mohsenipour, "Attitude Control of a Quadrotor with Optimized PID Controller," *Intell. Control Autom.*, vol. 2013, no. August, pp. 335–342, 2013.

[12] S. Bouabdallah, "Design and Control of Quadrotors With Application To Autonomous Flying," *Techniques*, vol. 3727, no. 3727, p. 61, 2007.

[13] Raju, M., & Scienc, A. (2010). *Dynamics Modeling and control of A Quad-Rotor Helicopter*. Memorial University of ewfoundland.

[14] Yongqiang Bai; Hao Liu; Zongying Shi; Yisheng Zhong, "Robust control of quadrotor unmanned air vehicles," *Control Conference (CCC), 2012 31st Chinese* , vol., no., pp.4462,4467, 25-27 July 2012

[15] EMRECANSU İÇMEZ. (2014). *Trajectory Tracking of a Quadrotor Unmanned Aerial Vehicle (UAV) VIA Attitude and position control.* MIDDLEEASTTECHNICALUNIVERSITY.

[16] S. H. Jeong and S. Jung, "A quad-rotor system for driving and flying missions by tilting mechanism of rotors: From design to control," *Mechatronics*, vol. 24, no. 8, pp. 1178–1188, 2014.

[17] a. L. L. Salih, M. Moghavvemi, H. a. F. a. F. Mohamed, and K. S. S. Gaeid, "Modelling and PID controller design for a quadrotor unmanned air vehicle," *Autom. Qual. Test. Robot. (AQTR), 2010 IEEE Int. Conf.*, vol. 1, pp. 1–5, 2010.

[18] A. Rodić and G. Mester, "The modeling and simulation of an autonomous quad-rotor microcopter in a virtual outdoor scenario," *Acta Polytech. Hungarica*, vol. 8, no. 4, pp. 107–122, 2011.

[19] García Carrillo, L. R., Dzul López, A. E., Lozano, R., Pégard, C., & SpringerLink (Online service). (2013). Quad Rotorcraft Control Vision-Based Hovering and Navigation. *Advances in Industrial Control*, XIX, 179 p. 117 illus., 74 illus. in color. http://doi.org/10.1007/978-1-4471-4399-4

[20] Deif, T., Kassem, A., El Baioumi, G., Modeling and Attitude Stabilization of Indoor Quad Rotor, (2014) *International Review of Aerospace Engineering (IREASE), 7*(2), pp. 43-47.

[21] P. Getsov, S. Zabunov, and G. Mardirossian, "Quad-Rotor Unmanned Helicopter Designs," vol. 3, no. September, pp. 77–82, 2014.

[22] P. C. Zambrano-robledo, "Simplifying quadrotor controllers by using simplified design models," pp. 4236–4241, 2013.

[23] G. B. Kim, T. K. Nguyen, A. Budiyono, J. K. Park, K. J. Yoon, and J. Shin, "Design and development of a class of rotorcraft-based UAV," *Int. J. Adv. Robot. Syst.*, vol. 10, 2013.

[24] P. E. I. Pounds, "Design, Construction and Control of a Large Quadrotor Micro Air Vehicle," no. September, 2007.

[25] Srikanth, M.B.; Dydek, Z.T.; Annaswamy, A.M.; Lavretsky, E., "A robust environment for simulation and testing of adaptive control for mini-UAVs," *American Control Conference, 2009. ACC '09.* , vol., no., pp.5398,5403, 10-12 June 2009

[26] D. Lara, A. Sanchez, R. Lozano, and P. Castillo, "Real-time embedded control system for VTOL aircrafts: Application to stabilize a quad-rotor helicopter," *Proc. IEEE Int. Conf. Control Appl.*, pp. 2553–2558, 2006.

[27] Tanveer, M.H.; Hazry, D.; Ahmed, S.F.; Joyo, M.K.; Warsi, F.A.; Kamaruddin, H.; Razlan, Z.M.; Wan, K.; Shahriman, A.B., "NMPC-PID based control structure design for avoiding uncertainties in attitude and altitude tracking control of quad-rotor (UAV)," *Signal Processing & its Applications (CSPA), 2014 IEEE 10th International Colloquium on* , vol., no., pp.117,122, 7-9 March 2014

[28] V. Hrishikeshavan, J. Black, and I. Chopra, "Design and Testing of a Quad Shrouded Rotor Micro Air Vehicle in Hover," *53rd AIAA/ASME/ASCE/AHS/ASC Struct. Struct. Dyn. Mater. Conf.*, 2014.

[29] V. Hrishikeshavan, J. Black, and I. Chopra, "Design and Performance of a Quad-Shrouded Rotor Micro Air Vehicle," *J. Aircr.*, vol. 51, no. 3, pp. 1–13, 2014.

[30] K. N. Shah, B. J. Dutt, and H. Modh, "' Quadrotor – An Unmanned Aerial Vehicle ,'" vol. 2, no. 1, pp. 1299–1303, 2014.

[31] How, J. P., Supervisor, T., & Modiano, E. H. (2012). *Design and Control of an Autonomous Variable-Pitch Quadrotor Helicopter*. MASSACHUSETTS INSTITUTE OF TECHNOLOGY.

[32] A. U. Batmaz, O. Elbir, and C. Kasnakoglu, "Design of a Quadrotor Roll Controller Using System Identification to Improve Empirical Results," *Int. J. Mater. Mech. Manuf.*, vol. 1, no. 4, pp. 347–349, 2013.

[33] Abeywardena, D.M.W.; Munasinghe, S.R., "Performance analysis of a Kalman Filter based attitude estimator for a Quad Rotor UAV," *Ultra Modern Telecommunications and Control Systems and Workshops (ICUMT), 2010 International Congress on* , vol., no., pp.466,471, 18-20 Oct. 2010

[34] M. Hassan Tanveer, S. Faiz Ahmed, D. Hazry, F. a. Warsi, and M. Kamran Joyo, "Stabilized controller design for attitude and altitude controlling of quad-rotor under disturbance and noisy conditions," *Am. J. Appl. Sci.*, vol. 10, no. 8, pp. 819–831, 2013.

[35] Qiang Jiang; Yong Zeng; Qiang Liu; Hua Jing, "Attitude and Heading Reference System for Quadrotor Based on MEMS Sensors," *Instrumentation, Measurement, Computer, Communication and Control (IMCCC), 2012 Second International Conference on* , vol., no., pp.1090,1093, 8-10 Dec. 2012

[36] Seungho Jeong; Seul Jung, "Vision-based localization of a quad-rotor system," *Ubiquitous Robots and Ambient Intelligence (URAI), 2012 9th International Conference on* , vol., no., pp.636,638, 26-28 Nov. 2012

[37] S. C. Quebe, "Modeling , Parameter Estimation , and Navigation of Indoor Quadrotor Robots," 2013.

[38] F. Q. Elizabeth and L. R. Garc, "2012 American Control Conference Quad-Rotor Switching Control : An Application for the Task of Path Following," pp. 4637–4642, 2012.

[39] Wei Guo, Joseph Horn, "Modeling and Simulation for the Development of a Quad-Rotor UAV Capable of Indoor Flight

[40] ," AIAA Modeling and Simulation Technologies Conference and Exhibit, 2006.

[41] Jialiang Wang, Hai Zhao, Yuanguo Bi, Xingchi Chen, Ruofan Zeng, and Shiliang Shao, "Quad-Rotor Helicopter Autonomous Navigation Based on Vanishing Point Algorithm," Journal of Applied Mathematics, vol. 2014, Article ID 567057, 12 pages, 2014. doi:10.1155/2014/567057

[42] Armen Mkrtchyan, Richard Schultz,"Vision-Based Autopilot Implementation Using a Quad-Rotor Helicopter," AIAA Infotech@Aerospace Conference, 2009.

[43] G. P. Tournier, "Six Degree of Freedom Estimation Using Monocular Vision and Moir ´ e Patterns by," 2006.

[44] G. Tournier, M. Valenti, J. How, and E. Feron, "Estimation and control of a quadrotor vehicle using monocular vision and moire patterns," ... , *Navig. Control ...* , no. August, pp. 1–16, 2006.

[45] A. R. Patel, M. a Patel, and D. R. Vyas, "Modeling and analysis of quadrotor using sliding mode control," *Proc. 2012 44th Southeast. Symp. Syst. Theory*, pp. 111–114, 2012.

[46] Alaeddin Bani Milhim, Youmin Zhang, Camille-Alain Rabbath, "Quad-Rotor UAV: High-Fidelity Modeling and Nonlinear PID Control," Guidance, Navigation, and Control and Co-located Conferences, 2010.

[47] Khatoon, S.; Gupta, D.; Das, L.K., "PID & LQR control for a quadrotor: Modeling and simulation," *Advances in Computing, Communications and Informatics (ICACCI, 2014 International Conference on* , vol., no., pp.796,802, 24-27 Sept. 2014

[48] Cavalcante Sa, R.; De Araujo, A.L.C.; Varela, A.T.; De A Barreto, G., "Construction and PID Control for Stability of an Unmanned Aerial Vehicle of the Type Quadrotor," *Robotics Symposium and Competition (LARS/LARC), 2013 Latin American* , vol., no., pp.95,99, 21-27 Oct. 2013

[49] G. Szafranski and R. Czyba, "Different Approaches of PID Control UAV Type Quadrotor," pp. 70–75, 2011.

[50] X. Li, S. Qu, and L. Cai, "Studing of PID Parameter Optimization Method for Attitude Controller of Quadrotor Simplified Kinetics Model of Control System," vol. 18, no. 2012, pp. 7805–7812, 2014.

[51] E. Abbasi and M. Mahjoob, and R. Yazdanpanah, "Controlling of Quadrotor UAV Using a Fuzzy System for Tuning the PID Gains in Hovering Mode," *10th Int. Conf. Adv. Comput. Technol.*, pp. 1–6, 2013.

[52] Pipatpaibul, P., & Ouyang, P. R. (2013). Application of Online Iterative Learning Tracking Control for Quadrotor UAVs. *ISRN Robotics, 2013*(Ilc), 2010. http://doi.org/10.5402/2013/476153

[53] M. J. Mohammed and A. A. Ali, "Design Optimal PID Controller for Quad Rotor System," vol. 106, no. 3, pp. 15–20, 2014.

[54] C. Algorithms, "A Review of Control Algorithms," *Library (Lond).*, no. December, pp. 547–556, 2009.

[55] M. Young, The Technical Writer's Handbook. Mill Valley, CA: University Science, 1989.

AUTHORS PROFILE

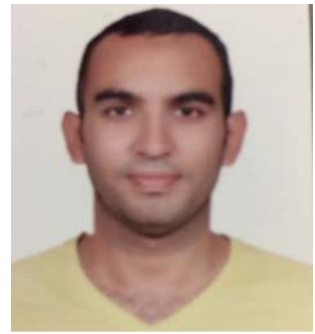

Tarek N.Dief

Bachelor,and Master (Cairo University), Ph.D.(continuing)

Teaching assitant in Aerospace Dep., Cairo University

Research Area: System Dynamics
Tarek.na3em@riam.kyushu-u.ac.jp

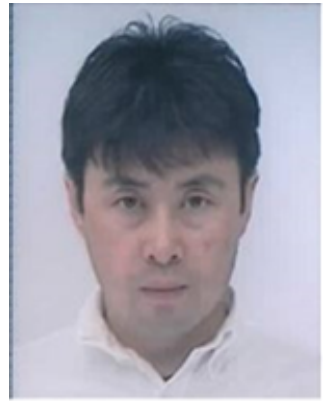

Shigeo Yoshida
Dr. Eng
Professor, Research Institute for Applied Mechanics
Kasuga, Fukuoka, Japan
Research Area: Wind Energy
yoshidas@riam.kyushu-u.ac.jp